**DEEGITS: Deep Learning based Framework for Measuring Heterogenous Traffic State in Challenging Traffic Scenarios**

**Muttahirul Islam**
Graduate Research Assistant, Department of Civil Engineering
Bangladesh University of Engineering and Technology (BUET), Dhaka-1000
Tel: 88-02-55167100 Ext. 7225, Fax: 880-2-58613046; *Email:muttahirulislamce@gmail.com*

**Nazmul Haque**
Lecturer, Accident Research Institute,
Bangladesh University of Engineering and Technology (BUET), Dhaka-1000
Tel: 88-02-55167100 Ext. 7897, Fax: 880-2-58613046; *Email:nhaque@ari.buet.ac.bd*

**Dr. Md. Hadiuzzaman, Corresponding Author**
Professor, Department of Civil Engineering
Bangladesh University of Engineering and Technology (BUET), Dhaka-1000
Tel: 88-02-55167100 Ext. 7225, Fax: 880-2-58613046; *Email: mhadiuzzaman@ce.buet.ac.bd*

## ABSTRACT

This paper presents DEEGITS (Deep Learning Based Heterogeneous Traffic State Measurement), a comprehensive framework that leverages state-of-the-art convolutional neural network (CNN) techniques to detect vehicles and pedestrian accurately and rapidly, as well as to measure traffic state in challenging scenarios (i.e., congestion, occlusion). In this study, we enrich the training dataset using a data fusion technique that enables simultaneous detection of vehicles and pedestrians. Afterward, image preprocessing and augmentation are performed to enhance the quality and quantity of the dataset. Transfer learning is employed on YOLOv8 pre-trained model to enhance the model's ability to identify diverse array of vehicles. Suitable hyperparameters are obtained using the Grid Search algorithm and the Stochastic Gradient Descent (SGD) optimizer outperforms other optimizers with these hyperparameters. Extensive experimentation and evaluation demonstrate significant accuracy rates in the detection framework, with the model achieving 0.794 mAP@0.5 in validation set and 0.786 mAP@0.5 in test set, surpassing previous benchmarks on similar datasets. DeepSORT multi-object tracking algorithm is adopted to track the detected vehicles and pedestrian in this study. Ultimately the framework is tested to measure heterogeneous traffic states in mixed traffic condition. Two locations with different traffic compositions and different congestion levels are selected: one is motorized dominant with moderate density, and the other is non-motorized dominant with higher density. The errors are statistically insignificant for both cases, with a correlation of 0.99-0.88 and 0.91-0.97 for measuring heterogeneous traffic flow and speed, respectively.

## INTRODUCTION

Accurate vehicles and pedestrian detection with tracking is crucial for efficient traffic flow measurement, real-time incident detection, and improved Intelligent Transportation Systems (ITS). Identifying vehicles and pedestrian enables the estimation of fundamental diagram parameters, such as jam density, capacity, and free flow speed, necessary for traffic flow simulation models and effective control and management strategies (1). It also has the utmost importance in road safety as it facilitates the automation of traffic conflict assessments, leading to proactive measures and improved overall road safety (2).

Various detectors, including loop detectors, LiDAR sensors, microwave detectors, GPS devices, and floating cars, are used to collect traffic flow information. However, their limitations in providing precise and comprehensive data, such as undetectable areas, connectivity issues, and high maintenance costs, highlight the importance of surveillance cameras for more accurate and cost-effective traffic monitoring and congestion alleviation (3, 4).

Traditional methods for obtaining traffic information from videos or images involve vehicle detection and counting using various image processing algorithms. These methods, such as Speeded Up Robust Features (SURF), background subtraction, and temporal difference, suffer from poor performance and limited accessibility in congested and complex traffic conditions or low frame rates (5). Before Convolutional Neural Networks (CNNs) emerged, feature-based object detection methods were widely used. One popular approach is optical flow, which calculates displacement vectors by searching for consecutive pixel matches and is often employed for road user classification (6).

Other conventional feature-based methods include Histogram of Oriented Gradients (HOG), Haar-like features and Scale-Invariant Feature Transform (SIFT). Although they compute abstractions of image information and classify road users based on specific features, they have difficulty in recognizing objects with repetitive patterns or textures, few distinctive features, and complex shapes or structures (7, 8). Their reliance on handcrafted features limits adaptability and hinders accurate vehicle detection and classification in real-world traffic scenarios. They also struggle with congestion, scale variations, occlusions, and lighting conditions (9).

In contrast, deep learning-based models, such as convolutional neural networks (CNNs), learn representations directly from the data, allowing them to capture complex patterns and adapt to different conditions more effectively (10). This ability to automatically learn features and hierarchies of representations contributes to the superior performance of deep learning-based models in road users' detection tasks.

The two-stage detection approach in CNNs involves using one model to extract object regions and another model to classify and refine the localization of the objects. R-CNN model pioneered the structure, followed by Fast R-CNN, Faster R-CNN, Mask R-CNN, MS-CNN and R-FCN, which improved accuracy and efficiency (11). These models excel in localizing and recognizing objects but at the expense of reduced inference speed. Challenges arise with scale variations and small objects, where region proposal struggles to accurately localize differently sized objects.

One-stage Convolutional Neural Networks (CNNs) overcome the limitations of two-stage models for object detection. They directly predict bounding boxes and class probabilities, eliminating the need for region proposals. This simplifies the model and improves speed, making it suitable for real-time applications. Single-stage models have evolved, with notable advancements such as SSD (12), YOLO (13), EfficientDet (14) and CenterNet (15).

Fan et al. (16) analyzed and optimized Faster R-CNN for vehicle detection, demonstrating the impact of parameter tuning and algorithmic modifications and achieved 71.22% accuracy on the subset KITTI dataset that only includes cars. Sochor et al. (17) introduced a fine-grained vehicle recognition system for traffic surveillance that automatically constructs 3D bounding boxes around vehicles without relying on 3D vehicle models. However, their proposed method struggles when multiple vehicles are in one image. Liang et al. (18) introduced a cascaded convolutional neural network (CNN) model for car detection and classification, where they focused solely on detecting cars and did not consider the detection and classification of other types of vehicles.

YOLO's real-time object detection capabilities have proven invaluable across various domains. YOLO models have been utilized in agriculture for crop detection and classification (*19*), and in the medical field for pill identification (*20*). Li et al. (*21*) employed YOLOv4 to detect passenger flow in subway stations, achieving better accuracy compared to other classical models. Bin Zuraimi and Kamaru Zaman (*22*) used YOLOv4 to detect four classes (Car, motorcycle, bus, truck) and recommended it is faster than the YOLOv3 model.

The tracking algorithm must be integrated with the detection model to achieve object trajectory. SORT (*23*), DeepSORT (*24*), and ByteTrack (*25*) are commonly used algorithms for object tracking. SORT utilizes object detection, Kalman filtering, and the Hungarian algorithm for efficient data association, accurately tracking objects. DeepSORT enhances SORT by incorporating appearance features and deep learning, improving accuracy and handling occlusion and ID switching in crowded scenes. ByteTrack is a state-of-the-art method that improves object trajectory accuracy by matching all detection boxes, regardless of scores. However, DeepSORT's accuracy in crowded environments depends on the detection model's accuracy. Integration of DeepSORT with the latest YOLO version has yet to be validated.

Open Dataset annotated with bounding box information is compulsory for training the YOLO model. The YOLO model is pre-trained on the MS COCO dataset (*26*), a widely used object detection and segmentation benchmark. It includes vehicle classes like cars, trucks, buses, motorcycles, and bicycles. However, it has insufficient variation, lack of fine-grained annotations, limited temporal information, and geographical bias. The CityPersons dataset (*27*) is widely used for pedestrian detection in urban areas. It offers annotated images focusing on pedestrians, but no vehicle classes are annotated. So, there needs to be a generalized dataset where pedestrians and all kinds of vehicles are labeled.

Existing deep learning methods are limited to detecting pedestrians or a few vehicles, neglecting local peculiarities. Researchers rarely attempt to detect both simultaneously, and no studies focus on detecting vehicles dominated in heterogenous conditions. Thus, there needs to be a generalized detection model to detect vehicles and pedestrians, accurately.

In this study, we present a comprehensive framework that leverages state-of-the-art convolutional neural network (CNN) techniques for classified detection of pedestrian and vehicles accurately and rapidly and measuring traffic state in challenging mixed traffic scenarios. The complementary dataset fusion technique is employed to prepare the desired dataset. Images are preprocessed, including fixing orientation and applying histogram equalization. The training dataset is augmented using various methods such as grid dropout, Gaussian noise, and mosaic to enhance detection effectiveness, even when vehicles are mostly occluded. Transfer learning is employed on YOLOv8 pre-trained model to enhance the model's ability to identify road users (vehicles and pedestrian) regardless of the geographical setting. Stochastic Gradient Descent (SGD) optimization are used to get the best results and to optimize other hyperparameters using Grid Search Algorithm. DeepSORT tracking algorithm has been integrated to obtain road users trajectories. Finally, this framework has been used to measure the traffic states (flow and speed) from the field and validate the result against the ground truth value.

The remainder of this paper has been partitioned as follows: Section 2 presents the methodology of detection model, tracking algorithm and measurement of traffic state using DEEGITS; Section 3 shows data preparation, analysis and performance of DEEGITS in field; finally, concluding remarks and future research scopes are given in Section 4.

## METHODOLOGY

The methodology of this research is divided into three components: (1) Detection and Classification Model; (2) Tracking Algorithm; (3) Traffic State Measurement; **Figure 1** shows the flow chart illustrating the connectivity among these components.

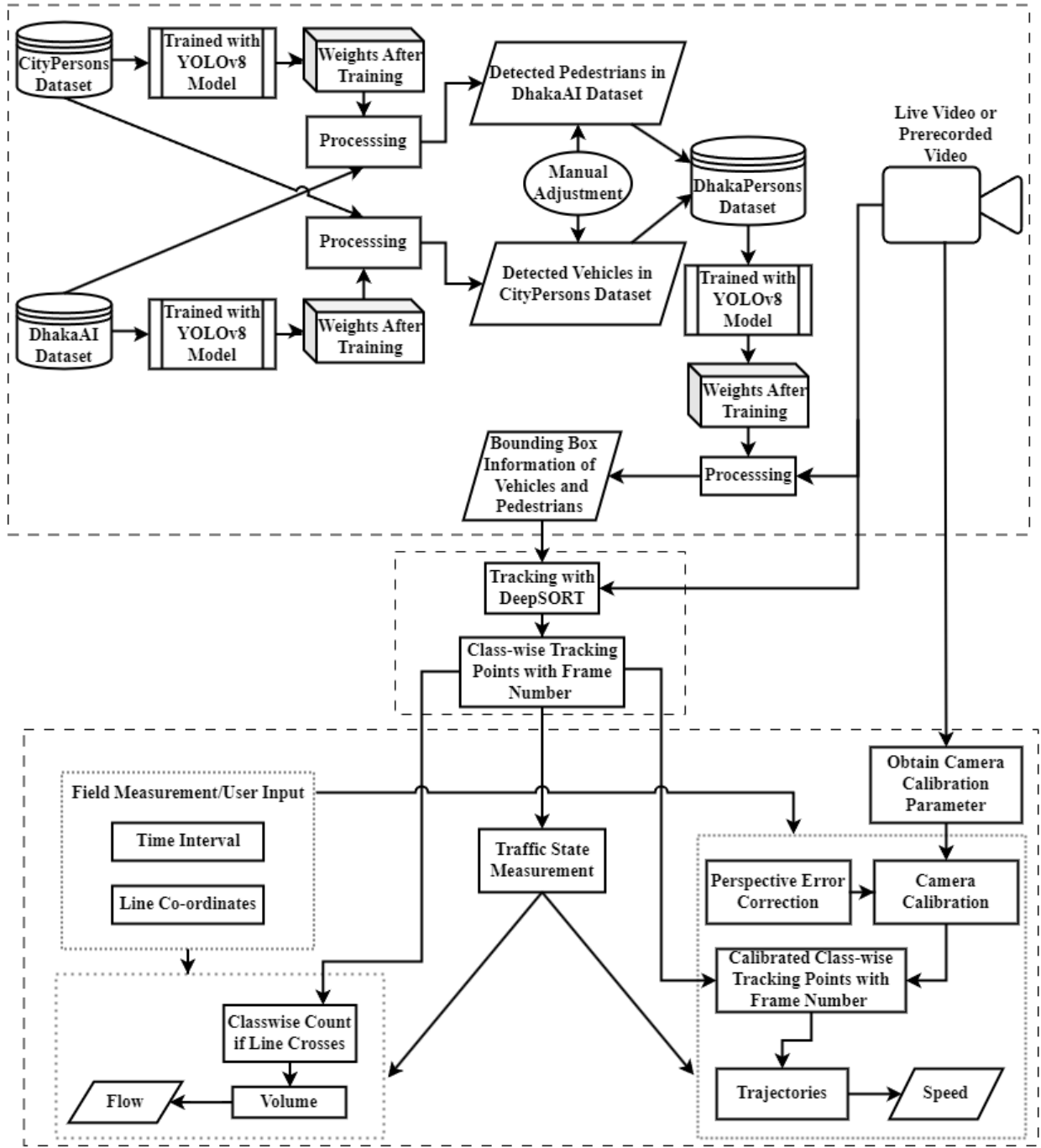

**FIGURE 1 Flow chart showing the architecture of Deep Learning Based Heterogeneous Traffic State Measurement (DEEGITS)**

**Detection and Classification Model**

YOLOv5 (*28*), a popular version of YOLO, has been widely used for vehicle and pedestrian detection, while YOLOv8 has seen limited adoption (*29*, *30*). However, YOLOv8 introduces significant improvements over YOLOv5, outperforming it in various aspects (*31*).

The notable changes in YOLOv8 include an anchor-free architecture, eliminating the need for anchor boxes and directly predicting object centers. This reduces the number of box predictions and speeds up post-processing steps like Non-Maximum Suppression. The head module has been updated to a decoupled structure, separating classification and detection heads and transitioning from Anchor-Based to Anchor-Free detection. The backbone network and neck module draw inspiration from YOLOv7's ELAN design with a modified C2f module (*32*). YOLOv8 uses Complete Intersection Over Union (CIoU) and Distribution Focal Loss (DFL) functions for bounding box loss and binary cross-entropy for classification loss. These losses have improved object detection performance, particularly when dealing with smaller objects. YOLOv8 retains YOLOv5's data augmentation but stops Mosaic augmentation in the final ten epochs. With its architectural improvements, YOLOv8 surpasses YOLOv5 in accuracy, achieving an average precision of 51.4% on the COCO dataset compared to YOLOv5's 50.5% (*31*).

**Tracking**

Road user's positions must be tracked in each video frame to determine their trajectories. Tracking relies on assigning unique identifications (IDs) to detections and maintaining them throughout the frames. Even if the object detector fails due to occlusion or overlap, the tracker can still predict and track the objects.

Integrating DeepSORT, a state-of-the-art multi-object tracking algorithm, with YOLOv8, IDs are assigned to tracked objects. Bounding box information from the detection model is used to develop motion and visual similarity search models. This estimates the state of the track in the next frame. A track has been defined in the state space using $(u, v, \gamma, h, \dot{x}, \dot{y}, \dot{\gamma}, \dot{h})$, where $(u, v)$ represents the center of the bounding box of the track, $\gamma$ represents the aspect ratio, $h$ represents the height of the box, and $(\dot{x}, \dot{y}, \dot{\gamma}, \dot{h})$ represents the velocity component of each state in image coordinates.

Two distances are calculated in the process. The association between the predicted Kalman states from the previous frame and the bounding box information arrived from the current frame is calculated by the squared Mahalanobis distance given in **Equation 1**. Here, the projection of $i$-th track distribution into measurement space is denoted by $(y_i, S_i)$ and $j$-th bounding box detection of new frame is expressed by $d_j$.

$$d^{(1)}(i, j) = (d_j - y_i)^T S_i^{-1} (d_j - y_i) \tag{1}$$

Unaccounted camera motion can introduce rapid displacements in the image plane, making the Mahalanobis distance a rather uninformed metric for tracking through occlusions. So, a second metric is used where for each bounding box detection $d_j$, an appearance descriptor $r_j$ with $\left\|r_j\right\| = 1$ is computed. Further, a gallery $R_k = \{r_k^{(i)}\}_{k=1}^{L_k}$ of the last $L_k = 100$ associated appearance descriptors for each track $k$ is kept. Then, the smallest cosine distance between the $i$-th track and $j$-th detection in appearance space is measured by using **Equation 2**.

$$d^{(2)}(i, j) = min\{1 - r_j * r_k^{(i)} | r_k^{(i)} \in R_i\} \tag{2}$$

A gate matrix is formed for taking decision whether the distances calculated from **Equation 1** and **2** are less than threshold value $t^{(1)}$ and $t^{(2)}$ respectively. These decisions are taken by using **Equation 3**. The association is referred as admissible if it is within the gating region of both matrices using **Equation 4.**

$$b_{(i,j)}^{(p)} = \begin{cases} 1, d^{(p)}(i, j) \leq t^{(p)} \\ 0, otherwise \end{cases}, \forall p = 1,2 \tag{3}$$

$$b_{i,j} = \prod_{m \in p} b_{i,j}^{(m)} \tag{4}$$

Finally cost matrix is formed using the weighted sum of previously calculated two distance matrices using **Equation 5**

$$c_{i,j} = \lambda d^{(1)}(i,j) + (1-\lambda)d^{(2)}(i,j) \tag{5}$$

The Hungarian algorithm optimizes the cost matrix for data association. It matches detections with previous tracks, creating "matched tracks." Unassociated detections and tracks undergo another matching process. "Matched tracks" are updated using the Kalman filter. "Unmatched tracks" inactive for three frames are removed, while "unmatched detections" creates tentative tracks and are deleted if unassociated within the first three frames. For more detailed information, readers are requested to see (*24*).

**Traffic State Measurement**

*Correction for Camera Skew*

Trajectory coordinates of each road users generated by the tracking algorithm are on skewed axis. Geometric correction is applied to those points using **Equations 6** and **7** for $Y$ and $X$ axes, respectively.

$$Y' = Y_0 + y \times \frac{\sin(\delta)}{\omega} \tag{6}$$

$$X' = X_0 + \frac{(x + \varphi \times \cot(\delta) \times Y)}{\varphi} \tag{7}$$

$$\varphi = \frac{R_X}{R'_x} \tag{8}$$

$$\omega = \frac{R_Y}{R'_Y} \tag{9}$$

Where,

$x, y$ = Uncalibrated trajectory coordinates along skewed $X$ axis and $Y$ axis respectively

$X_0, Y_0$ = Geodetic $X$ coordinate and $Y$ coordinate of the reference point of $(X', Y')$ coordinate system respectively

$X', Y'$ = Calibrated trajectory coordinates along Geodetic $X$ axis and $Y$ axis respectively

$\varphi, \omega$ = Magnification factor along $X$ axis and $Y$ axis respectively

$\delta$ = Angle of the skewed axis with $X$ axis

$R_X, R_Y$ = Length of a reference object along Geodetic $X$ axis and $Y$ axis respectively

$R'_x, R'_Y$ = Length of a reference object along skewed $X$ axis and $Y$ axis respectively

*Flow measurement*

Each detected vehicles and pedestrian have certain size represented by its bounding box information (centroid, width, and height). For simplicity, centroid point has been taken as simpler representation of detected vehicle or pedestrian. The measurement of each road user's position in space at tracked time frame $(t)$ constitutes its trajectory. $P_n$ represents the trajectory of $n$-th road users which is a zipped collection of all points in the desired time interval $T_i$

$$P_n = \left\{ \bigcup_{\forall t \in T_i} \left( X'(t), Y'(t) \right) \right\}, i = 1,2, \ldots \ldots N \tag{10}$$

Here $N$ is the number of time interval in total video duration. An input Line of Interest (LoI) is defined by the user for flow measurement. Then, an algorithm searches for the intersection of each road user's trajectory, $P_n$ with the LoI and increases its classified count value, $(S_n)^k$ if successes $(S_n = 1)$. The classified counts are shown after each interval $T_i$ and flow is measured by dividing the individual classified count with time interval where $k \in C$ , $C$ is categorical set of road user classes and $K$ is total road users identified of the class, $k$ in the time interval.

$$q_i^k = \frac{\sum_{j=1}^{K} s_j^k}{T_i} \tag{11}$$

*Speed Measurement*

For each time interval described before, Speed of individual tracked road users are calculated using **Equation 12.**

$$v(i)^k = \frac{\sum_{l=1}^{F} \sqrt{\left(X'(t_{l+1}) - X'(t_l)\right)^2 + \left(Y'(t_{l+1}) - Y'(t_l)\right)^2}}{\left(t_f - t_1\right) * f} \tag{12}$$

Where, $f$ is the video frame rate, $F$ is the maximum tracked frames and $t_{l+1}$ and $t_l$ are successive tracked frames in that time interval, $T_i$.

## DATA PREPARATION AND ANALYSIS

### Dataset for Training, Validating, and Testing of the Model

CityPersons dataset (*27*) focuses primarily on pedestrian detection. It is a subset of the Cityscapes dataset and annotates person instances only, which limits its utility for comprehensive mixed traffic analysis.

The MS COCO dataset (*26*) is a widely used benchmark for object detection tasks. It contains annotations for 80 categories, including various transportation-related classes such as persons, bicycles, cars, motorcycles, buses, trains, trucks, and boats. However, it lacks regional vehicle classes and may misclassify certain vehicles, such as 'Rickshaw' as 'Person'. In contrast, the DhakaAI dataset (*33*) focuses on vehicle classification and consists of 21 diverse classes, covering a wide array of vehicles. However, it lacks annotations for pedestrians in the dataset.

To overcome aforementioned limitations in existing datasets for detecting both vehicles and pedestrians in mixed traffic scenarios, a new dataset called "DhakaPersons" has been created by applying complementary data fusion techniques.

By training the DhakaAI dataset on the YOLOv8 model using transfer learning, mean average precision(mAP) of 0.755 in all classes has been achieved. Using this trained model, unannotated vehicles in the CityPersons have been detected. Similarly, the CityPersons dataset has been trained on YOLOv8 using transfer learning (mAP 0.687) to detect pedestrians in the DhakaAI dataset. Class-agnostic Non-Maximum Suppression has been applied to detect objects and bounding boxes, preventing multiple classes from being assigned to a single object. The mislabeled annotations are manually corrected and similar type vehicles are merged into single classes (i.e., cars and taxis into a single 'Private Passenger Car' class). A minimum of 5 pedestrian annotations is selected to ensure an adequate representation of pedestrians in each image.

The resulting 'DhakaPersons' dataset comprises 4576 images, divided into training (70%), validation (15%) and testing (15%) sets, consisting of 3200, 488 and 488 images, respectively. The dataset has 59576 annotated bounding boxes. The classes of the 'DhakaPersons' dataset are determined based on Bangladesh Road Transport Authority (BRTA) (*34*) registered vehicle classes to comply with field test addressing both motorized and non-motorized traffic . It should be mentioned that this classification is not confined to only BRTA, but it can also be transferred to other defined compositions. There are 14 classes, including Ambulance, Auto Rickshaw, Bicycle, Bus, Human Hauler, Microbus, Minibus, Motor Cycle, Pedestrian, Pickup, Private Passenger Car, Rickshaw, Special Purpose Vehicle, and Truck: instances (labeled objects) per class are 84, 3400, 913, 3681, 178, 2314, 436, 3032, 28947, 1225, 9951, 3537, 346 and 1532 respectively. 'Pedestrian' class has more than 28,000 instances annotated in the dataset. However, 'Ambulance', 'Special Purpose Vehicle' and 'Bicycle' are underrepresented.

**Image Augmentation and Preprocessing**

In the annotated dataset, some images are stored with varying orientations which mislead the training model. To address this, auto-orient has been processed on all the images in the dataset, ensuring consistent and accurate input for the model.

The dataset contains images of varying sizes, with 3,405 out of 4,576 training images having higher resolutions (>1024). The aspect ratio distribution reveals that 4,040 images have 2:1 aspect ratio. To ensure uniformity for training, all training images have been resized into 1024x512. Histogram Equalization is utilized to improve global contrast, particularly for images with close contrast values. 4% of background images are added to the dataset without labeling to improve model performance.

Augmentations have been applied to enhance the diversity and robustness of the data. Each training example has produced three outputs. A crop augmentation has been employed, ranging from a minimum zoom of 30% to a maximum zoom of 70%. Shear transformations have been applied horizontally and vertically, with a range of ±10°. Grayscale has been applied to 30% of the images to enhance their visual characteristics. Brightness and exposure adjustments have been made, varying between -20% and +20%. Cutout augmentations have been introduced with 15 boxes, each sized at 3%. The mosaic technique has been utilized to combine multiple images into a single training example. Bounding boxes have undergone a blur effect of up to 1px, ensuring smoother edges. After image augmentation, the training set contains 9674 images, including 50 background images.

**Model Training**

We have used 'yolov8s.pt' as pre-trained weight for the final training because it provides the best trade-off between inference speed and accuracy. For training the model, we utilized 1xTesla P100 GPU, which has 3584 CUDA cores and 16GB (16.28GB Usable) GDDR6 VRAM provided by Kaggle, a data science competition platform. The batch size is the number of samples that are processed at once during training. Larger batch sizes can lead to faster convergence but also require more memory. The batch size is adopted to 12 for this study, ensuring faster computing speed with less memory requirement. Three different epochs have been chosen: 50, 100 and 150; however, 100 epochs gave the best result. We optimized the training using Stochastic Gradient Descent (SGD), Adaptive Moment Estimation (Adam) and Root Mean Square Propagation (RMSProp) (35, 36). Although Adam has given the best validation accuracy but has failed to give satisfactory test accuracy and resulted in overfitting data.

On the other hand, SGD has similar validation and testing accuracy, removes the overfitting issue. The values for Momentum, learning and weight decay are adopted as 0.9, 0.01 and 0.0005, respectively, by the Grid Search algorithm. A 10% label smoothing process was introduced to improve the model's generalization ability during the training process. The Input image size is chosen to be 1280x1280 as the prepared training dataset by fusion contains small objects. The number of anchors is set to 3 and the number of classes is set to 14.

**Model Validation and Testing**

Three main evaluation metrics are used, including Precision, Recall, and F1-Score to determine the effectiveness of the trained model.

Precision quantifies the accuracy of positive predictions made by the model. It is calculated by **Equation 13.**

$$Precision = \frac{TP}{TP + FP} \tag{13}$$

Recall, also known as sensitivity, measures the model's ability to detect positive instances. It is calculated by **Equation 14**

$$Recall = \frac{TP}{TP + FN} \tag{14}$$

F1-Score (see **Equation 15**) is a weighted average of precision and recall, which takes into account both false positives and false negatives.

$$F1 - Score = 2 \times \frac{Precision \times Recall}{Precision + Recall}$$
(15)

Another commonly used metric for object detection models, mean Average Precision (mAP) has been also used to verify the model performance. The formula for calculating mean average precision is shown in **Equation 16** and **17**

$$AP = \sum_{m=0}^{m=r-1} [Recall(m) - Recall(m+1)] \times Precision(m)$$

$$mAP = \frac{1}{n} \sum_{k=1}^{k=n} AP_k$$
(17)

Where $r$ is the number of thresholds, $n$ is the number of classes and $AP_k$ is the average precision for class $k$. mAP summarizes the precision-recall curve into a single value, considering both false positives and false negatives. It balances the trade-offs between precision and recall, providing an overall evaluation of the model's performance in object detection tasks.

**Figure 2** displays the evaluation matrix for the validation dataset, while **Figure 3** presents the same for the test dataset.

The precision-confidence curve in **Figure 2(a)** demonstrates that all classes achieve a precision of 1 at a confidence threshold of 0.975, indicating high precision in predicting instances. **Figure 3(a)** shows a similar pattern, with all classes achieving a precision of 1 at a confidence threshold of 1.0. The recall-confidence curves in **Figures 2(b)** and **Figures 3(b)** reveal that the model successfully detects 91% and 92% of instances, respectively. Although the model exhibits good recall, it may miss a few instances of certain classes. However, the high precision observed in the precision-confidence curves ensures that the detected instances are likely true positives.

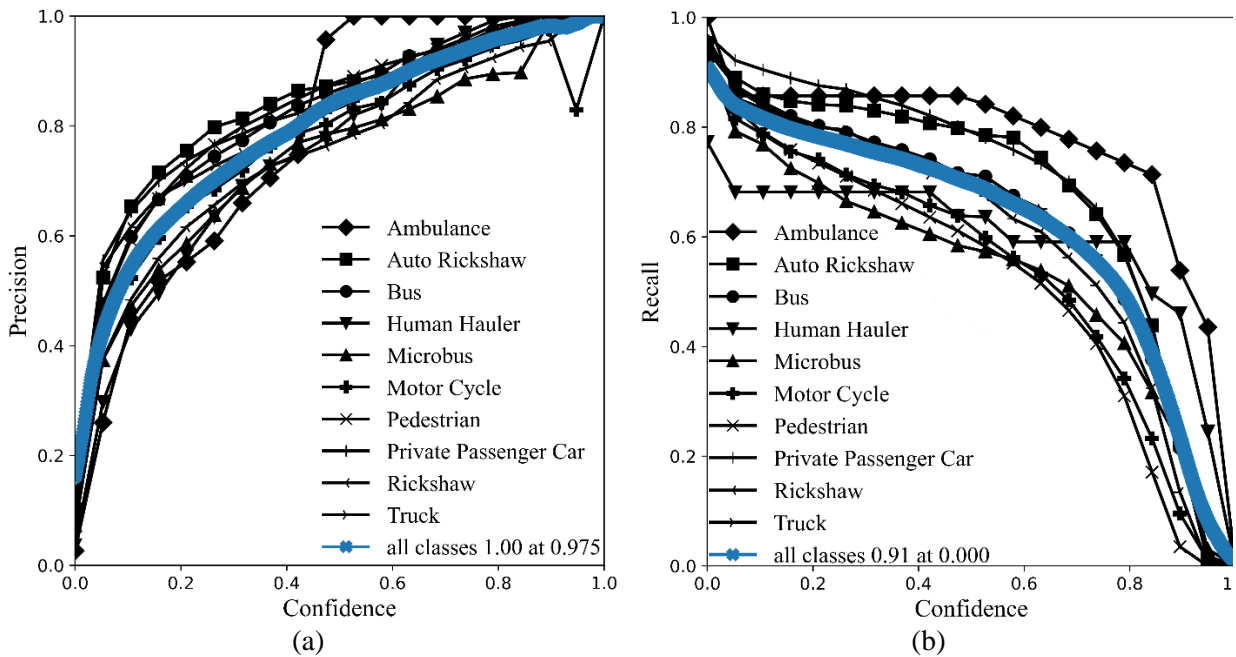

(a)                    (b)

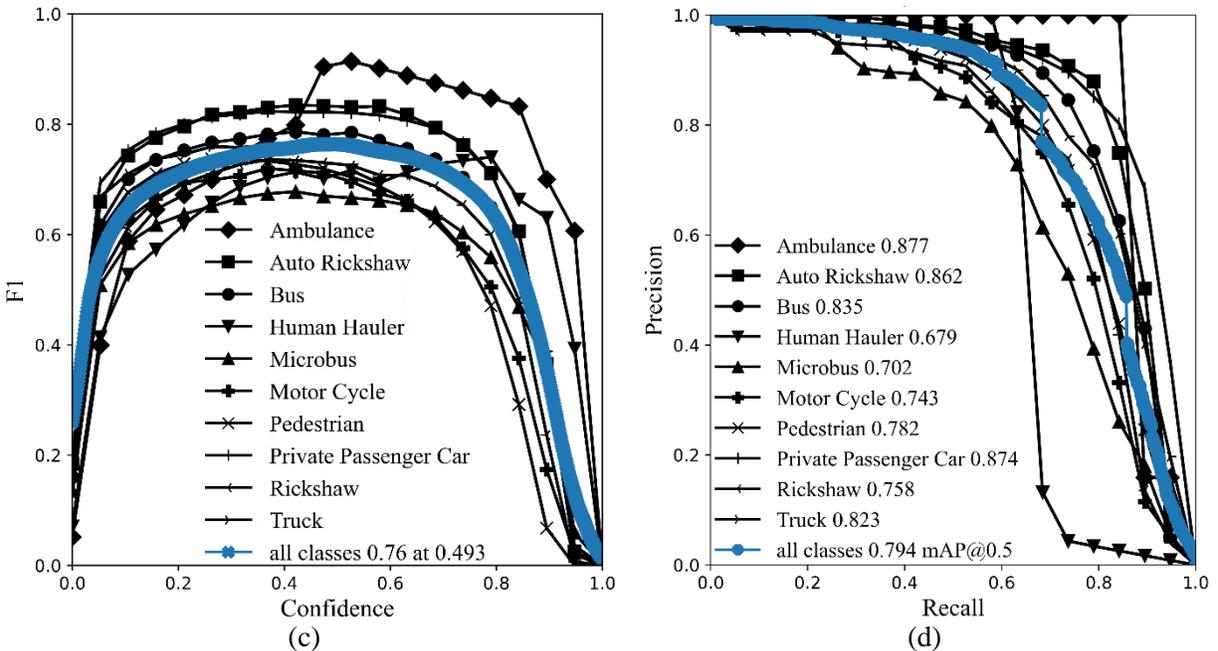

(c)           (d)

**FIGURE 2 Evaluation Metrices for Validation Set (a) Precision vs Confidence Curve (b) Recall vs Confidence Curve (c) F1 vs Confidence Curve (d) Precision vs Recall Curve**

The F1-confidence curve in **Figure 2(c)** indicates an F1 score of 0.76 at a confidence threshold of 0.493 in the validation dataset, meaning that the model correctly identifies the target class 76% of the time. **Figure 3(c)** shows an F1 score of 0.75 at a confidence threshold of 0.396, indicating a 75% correct identification rate. **Figure 2(d)** displays the model's performance on the validation dataset, with the 'Private Passenger Car' class achieving a precision of 0.874 and the 'Rickshaw' class achieving a precision of 0.758. In the test dataset (**Figure 3(d)**), the 'Auto Rickshaw' class exhibits the highest precision of 0.879, while the 'Bicycle' class has the lowest precision of 0.673. Overall, the model demonstrates good performance in pedestrian and vehicle detection.

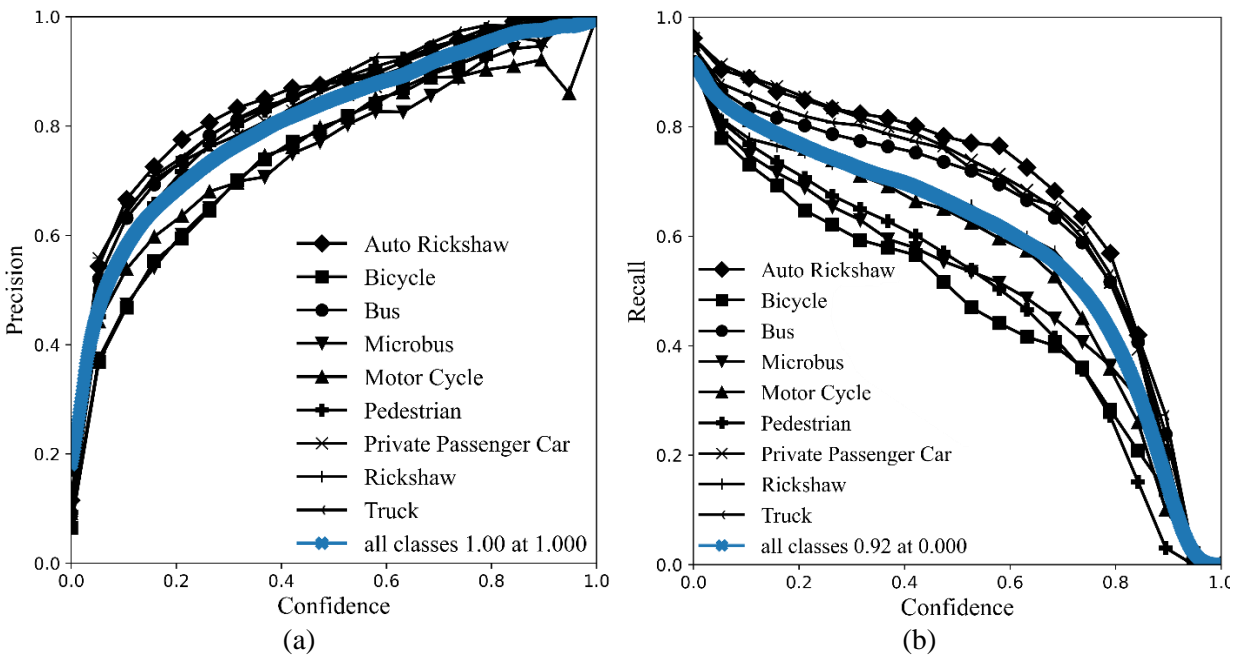

(a)           (b)

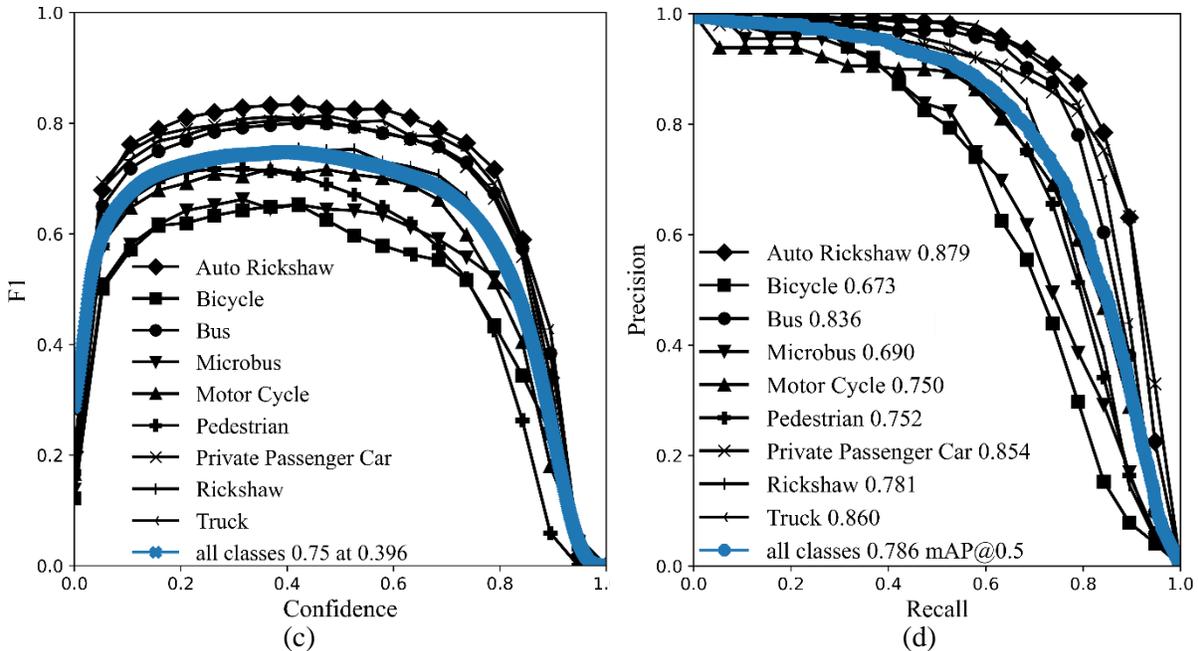

**Figure 3 Evaluation Metrices for Test Set (a) Precision vs Confidence Curve (b) Recall vs Confidence Curve (c) F1 vs Confidence Curve (d) Precision vs Recall Curve**

The small difference in mean average precision (mAP) between the validation and test sets (0.008) suggests that the model has overcome overfitting issues. Notably, the trained model surpasses the previous state-of-the-art accuracy in the DhakaAI dataset, achieving a mAP@0.5 of 0.786 compared to 0.458 (*37*).

**Table 1** presents the confusion matrix for pedestrian and vehicle classification on the DhakaPersons validation and test sets. The matrix consists of fourteen rows and columns, indicating true and predicted classes respectively. Diagonal elements represent correctly classified instances, while off-diagonal elements indicate misclassifications. The validation set exhibits high accuracy (>80%) for most classes, with 'Pedestrian' achieving 99% accuracy. Misclassifications include 'Special Purpose Vehicle' as 'Bicycle' and 'Ambulance' as 'Microbus'. Similarly, the test set shows accurate classifications for 'Auto Rickshaw' (98%) and 'Private Passenger Car' (94%), with misclassifications including 'Ambulance' as 'Microbus' and 'Bus'. The similarity in size between Microbus and Ambulance may contribute to the misclassifications in both sets. 'Special Purpose Vehicle' and 'Minibus' face challenges due to under-representation in the dataset.

## Table 1: Confusion Matrix for validation and test set

| | Ambulance | Auto Rickshaw | Bicycle | Bus | Human Hauler | Microbus | Minibus | Motor Cycle | Pedestrian | Pickup | Private Passenger Car | Rickshaw | Special Purpose Vehicle | Truck |
|---|---|---|---|---|---|---|---|---|---|---|---|---|---|---|
| **Ambulance** | **0.71** **(0.50)** | 0.00 (0.00) | 0.00 (0.00) | 0.00 (0.00) | 0.00 (0.00) | 0.00 (0.00) | 0.00 (0.00) | 0.00 (0.02) | 0.00 (0.00) | 0.01 (0.00) | 0.00 (0.00) | 0.00 (0.00) | 0.00 (0.00) | 0.00 (0.00) |
| **Auto Rickshaw** | 0.00 (0.00) | **0.96** **(0.98)** | 0.00 (0.00) | 0.00 (0.01) | 0.06 (0.06) | 0.01 (0.00) | 0.00 (0.00) | 0.01 (0.01) | 0.00 (0.00) | 0.02 (0.03) | 0.00 (0.00) | 0.01 (0.01) | 0.00 (0.04) | 0.00 (0.01) |
| **Bicycle** | 0.00 (0.00) | 0.00 (0.00) | **0.88** **(0.88)** | 0.00 (0.00) | 0.00 (0.00) | 0.01 (0.00) | 0.00 (0.00) | 0.01 (0.00) | 0.00 (0.00) | 0.00 (0.00) | 0.00 (0.00) | 0.00 (0.00) | 0.21 (0.17) | 0.00 (0.00) |
| **Bus** | 0.00 (0.13) | 0.00 (0.00) | 0.00 (0.00) | **0.97** **(0.97)** | 0.06 (0.00) | 0.00 (0.01) | 0.15 (0.08) | 0.00 (0.01) | 0.00 (0.00) | 0.04 (0.02) | 0.00 (0.00) | 0.00 (0.00) | 0.00 (0.00) | 0.05 (0.01) |
| **Human Hauler** | 0.00 (0.00) | 0.00 (0.00) | 0.00 (0.00) | 0.00 (0.00) | **0.78** **(0.75)** | 0.00 (0.00) | 0.00 (0.02) | 0.00 (0.00) | 0.00 (0.00) | 0.00 (0.01) | 0.00 (0.00) | 0.00 (0.00) | 0.04 (0.04) | 0.00 (0.00) |
| **Microbus** | 0.14 (0.38) | 0.01 (0.00) | 0.00 (0.01) | 0.00 (0.00) | 0.00 (0.06) | **0.80** **(0.79)** | 0.04 (0.03) | 0.01 (0.00) | 0.00 (0.00) | 0.02 (0.02) | 0.02 (0.00) | 0.00 (0.00) | 0.00 (0.04) | 0.00 (0.00) |
| **Minibus** | 0.00 (0.00) | 0.00 (0.00) | 0.00 (0.00) | 0.00 (0.00) | 0.00 (0.00) | 0.01 (0.00) | **0.52** **(0.65)** | 0.01 (0.01) | 0.00 (0.00) | 0.00 (0.00) | 0.00 (0.00) | 0.00 (0.00) | 0.00 (0.00) | 0.00 (0.00) |
| **Motor Cycle** | 0.00 (0.00) | 0.00 (0.00) | 0.02 (0.01) | 0.00 (0.00) | 0.00 (0.00) | 0.01 (0.01) | 0.13 (0.08) | **0.89** **(0.90)** | 0.00 (0.00) | 0.00 (0.00) | 0.01 (0.02) | 0.00 (0.00) | 0.00 (0.00) | 0.00 (0.00) |
| **Pedestrian** | 0.00 (0.00) | 0.00 (0.00) | 0.04 (0.06) | 0.00 (0.00) | 0.00 (0.00) | 0.00 (0.01) | 0.00 (0.00) | 0.01 (0.01) | **0.99** **(0.99)** | 0.00 (0.00) | 0.00 (0.00) | 0.01 (0.01) | 0.11 (0.08) | 0.00 (0.00) |
| **Pickup** | 0.14 (0.00) | 0.00 (0.00) | 0.00 (0.06) | 0.00 (0.00) | 0.06 (0.00) | 0.01 (0.01) | 0.00 (0.00) | 0.00 (0.00) | 0.00 (0.00) | **0.71** **(0.84)** | 0.00 (0.00) | 0.00 (0.00) | 0.00 (0.04) | 0.03 (0.07) |
| **Private Passenger Car** | 0.00 (0.00) | 0.00 (0.01) | 0.00 (0.00) | 0.01 (0.00) | 0.00 (0.00) | 0.13 (0.15) | 0.17 (0.13) | 0.06 (0.06) | 0.00 (0.00) | 0.06 (0.02) | **0.96** **(0.94)** | 0.00 (0.00) | 0.00 (0.00) | 0.00 (0.00) |
| **Rickshaw** | 0.00 (0.00) | 0.01 (0.01) | 0.03 (0.01) | 0.00 (0.00) | 0.06 (0.00) | 0.00 (0.15) | 0.00 (0.13) | 0.01 (0.00) | 0.00 (0.00) | 0.01 (0.01) | 0.00 (0.00) | **0.98** **(0.99)** | 0.00 (0.00) | 0.00 (0.00) |
| **Special Purpose Vehicle** | 0.00 (0.00) | 0.00 (0.00) | 0.02 (0.02) | 0.00 (0.00) | 0.00 (0.00) | 0.00 (0.06) | 0.00 (0.00) | 0.00 (0.00) | 0.00 (0.00) | 0.00 (0.01) | 0.00 (0.00) | 0.00 (0.00) | **0.64** **(0.50)** | 0.00 (0.00) |
| **Truck** | 0.00 (0.00) | 0.00 (0.00) | 0.00 (0.00) | 0.01 (0.00) | 0.00 (0.00) | 0.00 (0.00) | 0.00 (0.00) | 0.00 (0.00) | 0.00 (0.00) | 0.14 (0.05) | 0.00 (0.00) | 0.00 (0.00) | 0.00 (0.08) | **0.91** **(0.90)** |

* The values within the braces represents confusion matrix for the test set

**Field Validation**

Two locations in Dhaka city are chosen to validate the Traffic measurement methodology incorporated within DEEGITS. These locations comprise of different roadway and traffic characteristics. Specifically, the first location is a 4-legged urban signalized intersection containing heterogeneous motorized vehicles only. The intersection has three lanes in each leg. The dominant vehicle is Motorbike and the stream also contains a considerable number of side-facing buses, imposing a greater challenge to the framework to detect and consistently track smaller vehicles through occlusion. Another challenge is that the vehicles have long shadows and these shadows are falling upon other vehicles.

The next location is a 4-legged urban signalized intersection containing mixed motorized and non-motorized vehicles. The dominant vehicle is Rickshaw which is a non-motorized vehicle (NMV). The stream also contains a considerable number of bicycles. This location also imposes a challenging situation to detect, classify and track non-motorized traffic as a whole instead of separately along with a person sitting on the NMV. This allows for analyzing the accuracy of the data fusion technique. Although this video has smaller vehicle shadows, there exist shadows and occlusions of foreign objects (i.e., trees, electric poles).

Among these locations, traffic movements in both roadway segments show poor lane discipline. 60 minutes videos with 25 frame rates have been captured in each of the study areas. The mounting height of the cameras is at least 20ft and their angle is less than 45 degrees to reduce the detection of foreign objects (e.g., sky, birds etc.). For the same period, ground truth data (speed and flow) has also been collected from the video through manual post-processing. For ground truth and DEEGITS-based speed measurement, a strip of 88ft is chosen within the field of vision (FOV).

The camera calibration parameters set $\pi = \{\varphi, \omega, \delta, X_0, Y_0\}$ are measured to be $\pi_1 = \{53.9782, 55.5444, 233865.97, 2630553.14\}$ for the first location and $\pi_2 = \{60.0909, 46.0172, -78.5563, -76.1593, 234493.41, 2627654.18\}$ for the second location, respectively. The values of the $X_0$ and $Y_0$ are in Universal Transverse Mercator (UTM) 46N projected coordinate system.

Hyperparameters set obtained in the training step described in the sub-section 'Hyperparameter Tuning' yielded maximum precision-recall values and are used as required hyperparameters for the proposed framework. The framework analyzed the videos, and flow and speed measurements are obtained as these are the mostly used traffic state measurement variables (*38*, *39*). Afterwards, these measurements are compared with the ground truth both qualitatively (**Figure 4** and **5**) and quantitatively.

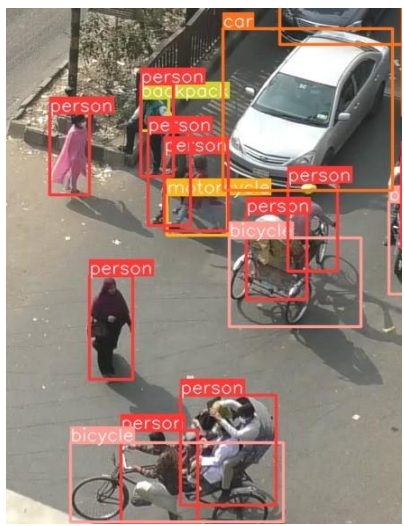 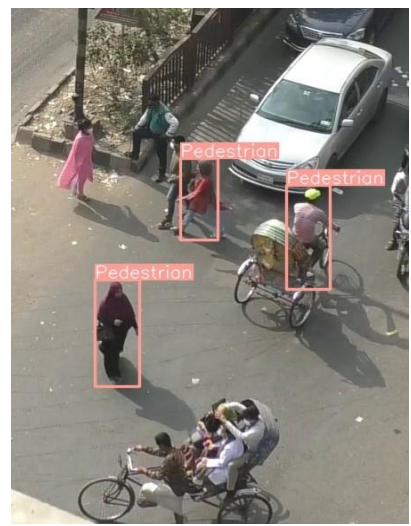

(a)                                     (b)

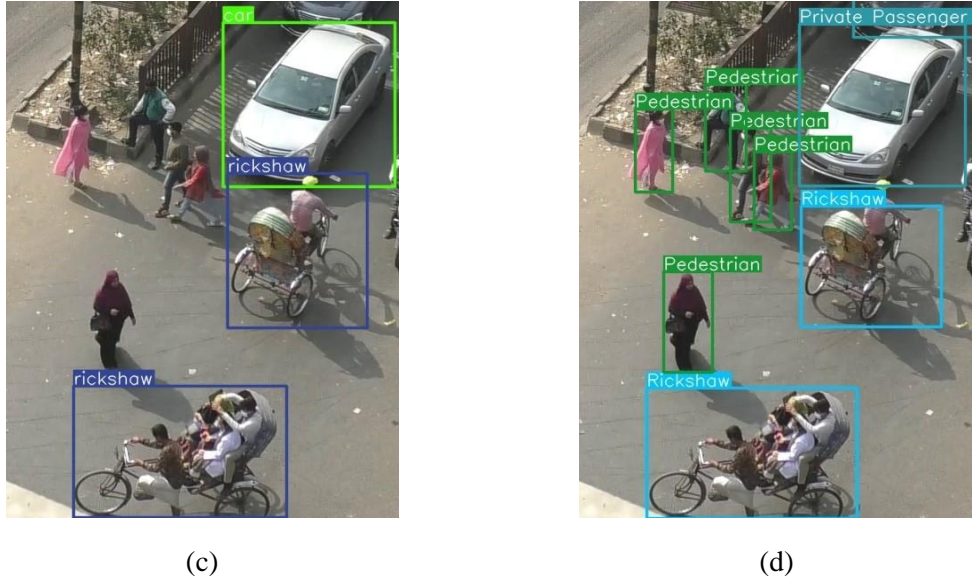

(c)                                    (d)

**Figure 4: Qualitative Evaluation of Vehicle and Pedestrian Identification** (a) Original YOLOv8 model detects car and person; however it misclassify rickshaw as bicycle and can't differ between pedestrian and passenger (b) YOLOv8 model trained with CityPersons dataset only detects some of the pedestrian and fails to detect any vehicle (c) YOLOv8 model trained with DhakaAI dataset, although correctly detected vehicles but misses pedestrian detection (d) Improved YOLOv8 model trained with 'DhakaPersons' dataset can accurately detect motorized and non-motorized vehicles along with pedestrian.

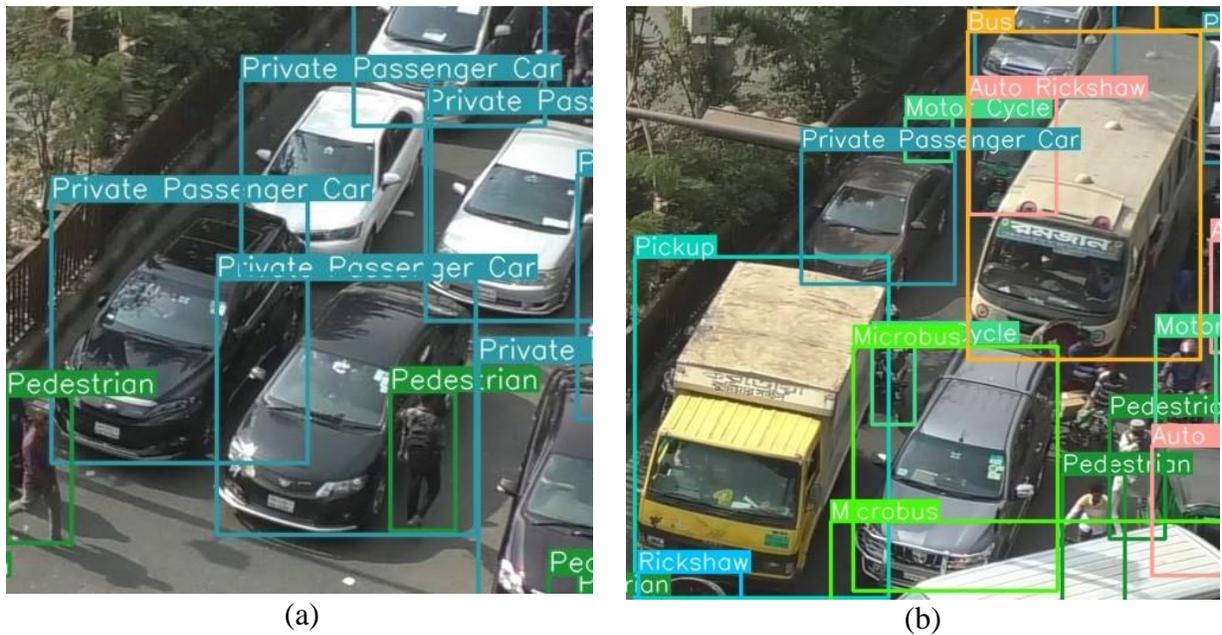

(a)                                    (b)

**Figure 5: Qualitative Evaluation of Model in Highly Congested Scenario** (a) The two black 'Private Passenger Car' has small lateral gap between each other whereas the white 'Private Passenger Car' is just behind of the front one. Trained YOLOv8 model can accurately detect closely gapped vehicles, both laterally and longitudinally in congested Scenario. (b) Small vehicle like 'Auto Rickshaw' and 'Motor Cycle' are partially occluded by large vehicle, 'Bus'. Trained YOLOv8 model shows its efficiency by accurately detecting those occluded vehicles

**Figure 6** and **7** show the performance of the framework for measuring flow and speed at different locations considered in this study. In that figure, ±10 vehicles and ±2.5 km/h error lines are established for vehicle count and speed validation, respectively (see **Figure 6(b)-(d)** and **Figure 7(b)-(d)**). Additionally, residual plots (see **Figure 6(a)-(c)** and **Figure 7(a)-(c)**) are incorporated to understand the sign of the error. RMSE values for the two study locations are 2.77 and 3.48 in measuring flow, whereas they are 1.78 and 1.61 in measuring speed.

Furthermore, a two-tailed t-test is performed to confirm whether there is any significant difference between the ground truth and the measurements provided by the framework. In this context, the null hypothesis assumes that the mean of ground truth and DEEGITS-based measurement is the same. The t-values for the two study locations are 1.804 and 1.973 in measuring flow, whereas 1.938 and 2.01 in measuring speed. The t-values for framework-based speed and flow measurements in all the locations are less than the t-critical (2.015) value at a 95% confidence level. Moreover, all the p-values are greater than 0.05. It results in accepting the null hypothesis. This indicates that the differences in the mean of ground truth and DEEGITS-based measurements are insignificant.

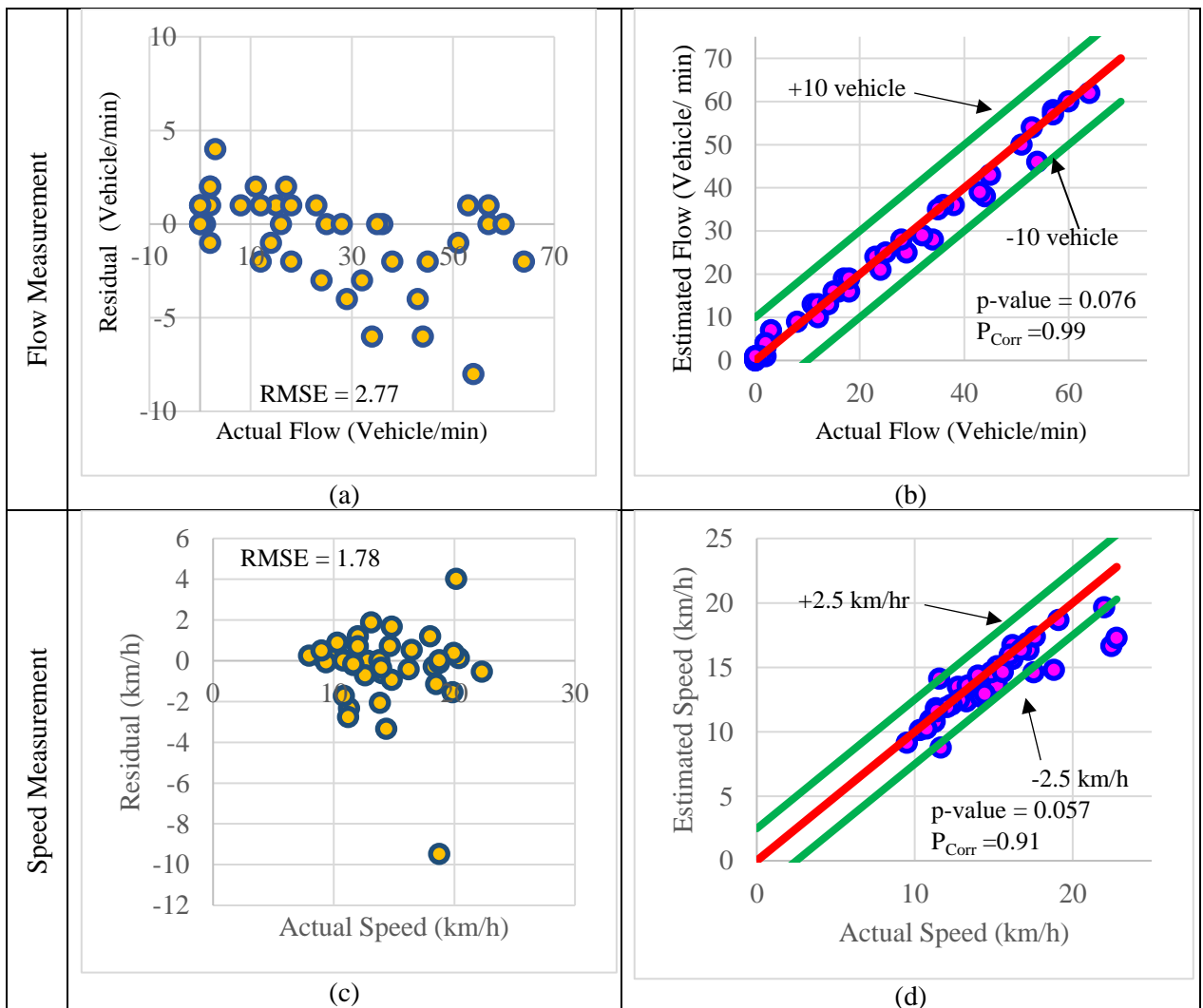

**Figure 6 Performance evaluation of DEEGITS-based flow and speed measurement for motorized vehicle dominant traffic stream ; (a)-(c) Residual Plot; and (b)-(d) Validation plot**

Pearson Correlations for the two study locations are 0.99 and 0.88 in measuring flow, whereas 0.91 and 0.97 in measuring speed. Hence, graphical and statistical analyses show that the framework produces better results in terms of flow measurement for the first location. The posterior investigation revealed that the second location contains several non-standard vehicles (i.e., push-carts, hawker vans) or vehicles with manual decoration (or, modification) that significantly mismatches the training dataset. In contrast, the speed measurement at the second location has been found to be better to that at the first location. This difference in performance is due to the presence of side-facing buses obstructing the view of smaller vehicles, causing occlusion issues at the first location. Such events cause interruptions in continuous tracking and loss in tracking points when the small vehicle occludes behind a large vehicle. Although the small vehicle emerges later, the preceding track points are lost, and the vehicle starts with a new tracking id with a new speed record keeping the previous one unfinished. Such a condition causes an error in the speed measurement in the first location.

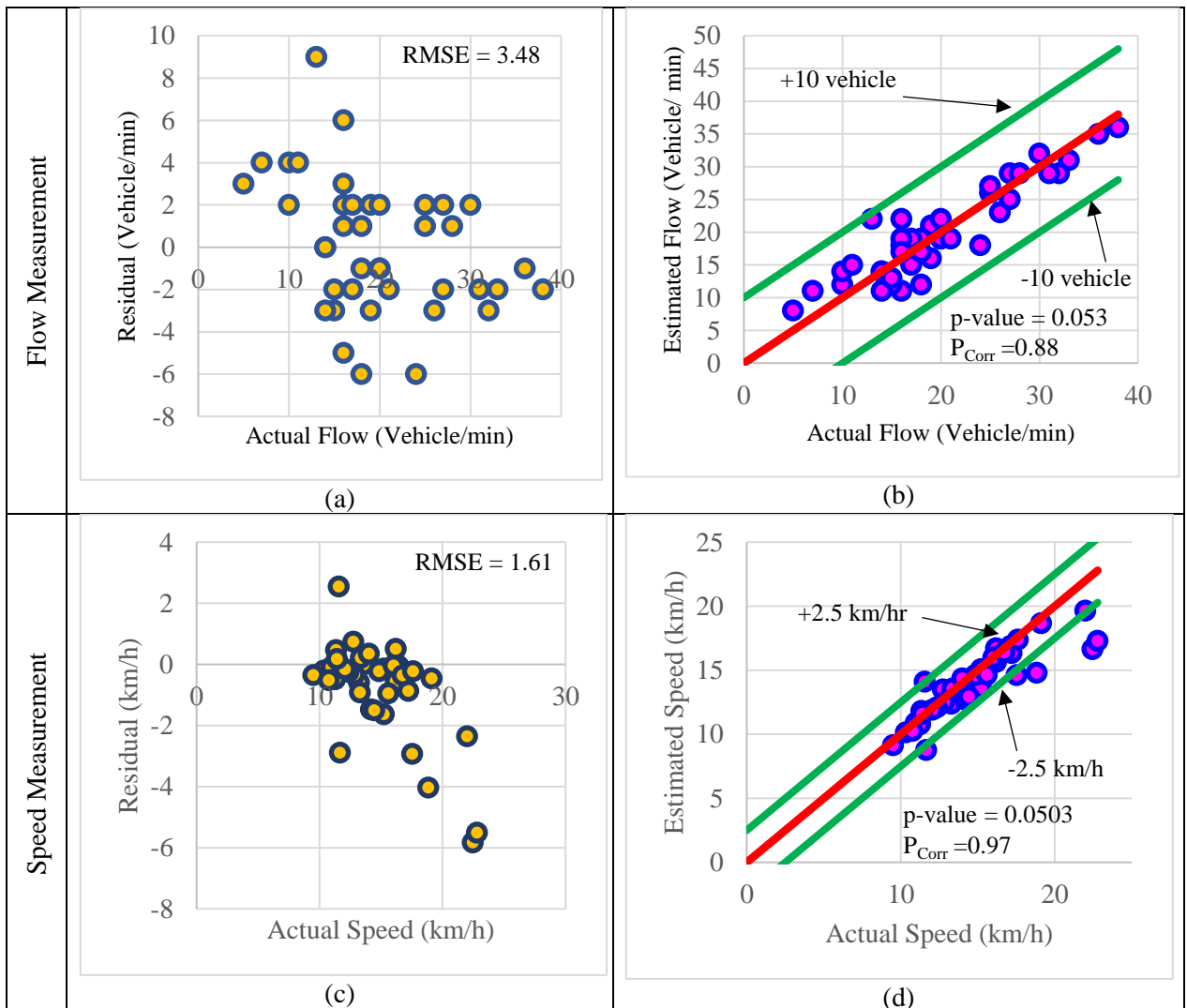

**Figure 7 Performance evaluation of DEEGITS-based flow and speed measurement for non-motorized vehicle dominant traffic stream; (a)-(c) Residual Plot; and (b)-(d) Validation plot**

## CONCLUDING REMARKS

This study presents a comprehensive framework that leverages state-of-the-art convolutional neural network (CNN) techniques to measure traffic states in congested mixed traffic scenarios. DEEGITS (Deep Learning Based Heterogeneous Traffic State Measurement), the robust framework which incorporates the proposed methodology, has been developed to facilitate simultaneous real-time classified traffic (i.e., vehicle and pedestrian) detection and heterogenous traffic state measurement. The training dataset has been enriched using a data fusion technique that simultaneously enables vehicle and pedestrian detection. Rigorous image preprocessing and augmentation are performed to enhance the quality and quantity of the dataset. Transfer learning is employed on YOLOv8 pre-trained model to enhance the naïve model's ability to identify local vehicles. Suitable hyperparameters have been obtained using the Grid Search algorithm and the Stochastic Gradient Descent (SGD) optimizer performs best with these hyperparameters. After extensive experimentation and evaluation, the detection framework exhibits remarkable accuracy rates. The model achieved a validation accuracy of 0.794 mAP@0.5 and a testing accuracy of 0.786 mAP@0.5, surpassing previous benchmarks on similar datasets. This framework has also provided accurate and informative traffic measurements under different road and traffic characteristics. Specifically, the framework provides more than 91% Pearson correlation values for speed measurement at all the study locations. In contrast, this value is around 88% in the case of flow measurement.

Moreover, among the study locations, analyses show that DEEGITS produces better results in the motorized traffic stream than the traffic stream with NMV. The main reason behind the lower correlation is the presence of non-standard and peculiar vehicles that are neither included in the training data nor has adequate sample number. On the other hand, speed measurement shows a reverse trend because the motorized traffic stream was suffering from severe occlusion beyond the framework's capacity. However, all errors are tested statistically and it found that all the errors are insignificant at a 95% confidence interval. All the above results show the robustness of the developed tool in measuring traffic flow and speed, which indicates vehicle detection accuracy. Notably, the dataset fusion technique ensures the simultaneous detection of vehicles and pedestrians. Preprocessing this dataset for training by fixing orientation and applying grayscale and histogram equalization removes any unnecessary noise from the dataset. Data augmentation techniques incorporating grid dropout, gaussian noise, and mosaic increase their effectiveness in detecting even when vehicles are in occluded condition. Moreover, transfer learning enhances the model's ability to identify local vehicles. Stochastic Gradient Descent (SGD) is incorporated to optimize hyperparameters, resulting in accurate vehicle classification without overfitting. Integration of the DeepSORT tracking algorithm enables the framework to detect pedestrians and vehicle trajectory, which ultimately resulted in the measurement of flow and speed.

DEEGITS, as a framework, incorporates different components utilizing different robust algorithms. It can save extracted real-time data for offline and online videos in a delimited text file (.txt) with versatile readability with an unrestricted storing facility (number of maximum columns and rows). At the same time, the data requires very less memory than other spreadsheet formats (.xlsx, .csv, .ods) due to its text format file structure. However, the prototype DEEGITS system developed in this study cannot handle severe vehicle occlusions and the presence of foreign vehicle types at the current stage. Depending on the presence and severity of these problems, traffic measurement accuracy may get affected. In the future, the authors plan to improve this framework by incorporating more vehicle classes and a large training dataset to remove its weaknesses, as stated above.

Moreover, the class balancing will be ensured to maintain uniformity among the instances of each vehicle class. Likewise, the current research scope does not include validating the framework at night. However, the YOLOv8 structure can also classify vehicles at night. Additionally, nighttime vehicle detection is another task that includes challenges other than daytime. The nighttime vehicle has partial visibility and glaring effect as critical challenges for this framework. Night vision infrared cameras can also pose different challenges by altering the actual color of the vehicle by converting the images into a greyscale that lacks multiple color channels. Thus, the framework keeps a scope to verify it at nighttime. Moreover, trajectory extraction, estimation, reconstruction, and prediction in the autonomous driving environment have become urgent in different research fields, like predicting traffic conflicts and anticipating crashes for vehicles and pedestrians. The developed framework welcomes the addition of any new component for such research initiatives.

## ACKNOWLEDGEMENT

This research work is financially supported by the research expense grant (Grant No. 69:3221109(16)) of Bangladesh University of Engineering and Technology (BUET). Their financial support has been played a critical role in the successful completion of this research.

## AUTHOR CONTRIBUTIONS

The authors confirm contribution to the paper as follows: study conception and design: M. Islam, N. Haque, M. Hadiuzzaman; data collection: M. Islam; analysis and interpretation of results: M. Islam, N. Haque; draft manuscript preparation: M. Islam. N. Haque. All authors reviewed the results and approved the final version of the manuscript.